\documentclass[10pt,twocolumn,letterpaper]{article}

\usepackage[pagenumbers]{cvpr} 

\usepackage{graphicx}
\usepackage{amsmath}
\usepackage{amssymb}
\usepackage{booktabs}
\usepackage{comment}
\usepackage{makecell}
\usepackage{multirow}
\usepackage{graphicx}
\usepackage[table,xcdraw]{xcolor}
\usepackage[normalem]{ulem}
\useunder{\uline}{\ul}{}
\usepackage[pagebackref,breaklinks,colorlinks]{hyperref}

\usepackage[capitalize]{cleveref}
\crefname{section}{Sec.}{Secs.}
\Crefname{section}{Section}{Sections}
\Crefname{table}{Table}{Tables}
\crefname{table}{Tab.}{Tabs.}

\usepackage{microtype} 

\def\cvprPaperID{59} 
\def\confName{CVPR}
\def\confYear{2023}

\begin{document}

\title{Attention Modules Improve Modern Image-Level Anomaly Detection: \\ A DifferNet Case Study}

\author{André Luiz B. Vieira e Silva$^{1}$
\and
Francisco Simões$^{1,3}$
\and
Danny Kowerko$^{2}$
\and
Tobias Schlosser$^{2}$
\and
Felipe Battisti$^{1}$
\and
Veronica Teichrieb$^{1}$
\and
$^{1}$ Voxar Labs, Centro de Informática, Universidade Federal de Pernambuco, Brazil\\
$^{2}$ Junior Professorship of Media Computing, Chemnitz University of Technology, Germany\\
$^{3}$ Visual Computing Lab, DC, Universidade Federal Rural de Pernambuco, Brazil\\
}
\maketitle

\begin{abstract}
    Within (semi-)automated visual inspection, learning-based approaches for assessing visual defects, including deep neural networks, enable the processing of otherwise small defect patterns in pixel size on high-resolution imagery. The emergence of these often rarely occurring defect patterns explains the general need for labeled data corpora. To not only alleviate this issue but to furthermore advance the current state of the art in unsupervised visual inspection, this contribution proposes a DifferNet-based solution enhanced with attention modules utilizing SENet and CBAM as backbone~-- AttentDifferNet~-- to improve the detection and classification capabilities on three different visual inspection and anomaly detection datasets: MVTec AD, InsPLAD-fault, and Semiconductor Wafer. In comparison to the current state of the art, it is shown that AttentDifferNet achieves improved results, which are, in turn, highlighted throughout our quantitative as well as qualitative evaluation, indicated by a general improvement in AUC of $\mathit{94.34}$ vs. $\mathit{92.46}$, $\mathit{96.67}$ vs. $\mathit{94.69}$, and $\mathit{90.20}$ vs. $\mathit{88.74}$~\%. As our variants to AttentDifferNet show great prospects in the context of currently investigated approaches, a baseline is formulated, emphasizing the importance of attention for anomaly detection.
\end{abstract}

\section{Introduction}
\label{sec:intro}

The automation of visual defect inspection can reduce inspection costs and security risks within real-world applications in the industry. Areas such as manufacturing, healthcare, or power delivery, often suffer from a scarcity of defective samples to train deep learning methods due to the high financial and social impact of defects \cite{bergmann2019mvtec, rudolph2021differnet, yu2021fastflow,schlegl2019fanogan,han2021madgan,akcay2019skipganomaly,roth2022patchcore,defard2021padim,ge2021anomaly}. This increases the importance of unsupervised anomaly detection methods \cite{liu2023deep}, which often rely on mostly normal or flawless samples during model training. They extract unique information from those samples, e.g., data distributions, whereby they can discriminate between flawless and anomalous samples during test time. 

Recent public datasets for anomaly detection, such as MVTec AD \cite{bergmann2019mvtec} and Magnetic Tiles Defects (MTD) \cite{huang2020surface}, fostered the proposition of new anomaly detection methods \cite{rudolph2022csflow, rudolph2021differnet, roth2022patchcore, yu2021fastflow, gudovskiy2022cflowad, defard2021padim}. The MVTec AD is the most used dataset for industrial anomaly detection. However, it only presents limited challenges from the manufacturing industry, i.e., components are captured under a controlled environment. On the contrary, anomaly detection in the wild, e.g., in power line inspections, is an open problem due to the lack of public datasets and benchmarks, for which additional challenges arise from often uncontrolled environments as changes in the background, lighting, scale, perspective, image resolution, object orientation, and occlusion.

Nowadays, a popular class of unsupervised approaches for anomaly detection is based on feature embedding manipulation \cite{liu2023deep}. Two techniques within this approach are distribution mapping through normalizing flows \cite{rudolph2022csflow, gudovskiy2022cflowad, rudolph2021differnet, yu2021fastflow} and feature memory bank \cite{defard2021padim, roth2022patchcore, cohen2020spade}.
Normalizing Flows are commonly used for density estimation and have become popular since they can model complex probability distributions using simpler ones, e.g., normal distributions \cite{rudolph2021differnet}. In a feature memory bank, the extracted features are stored in a memory bank, whereas each method uses a different approach to how the features are grouped and relate to each other.
 
To improve the spatial and/or channel encoding, multiple image-level anomaly detection methods apply attention mechanisms \cite{takimoto2022anomaly,song2023research,sun2022scale,ge2021anomaly}. In other words, they highlight relevant information from foreground objects while concealing the background and other less relevant image regions and objects \cite{ristea2022selfsupervised,bhattacharya2021interleaved}. On a similar path, in \cite{yan2022cainnflow}, the authors proposed to apply attention blocks during the normalizing flow step, which can lead to complex modifications due to their mathematically invertible nature. Two of the most popular ones are the Squeeze-and-Excitation network (SENet) \cite{hu2018senet} and the Convolutional Block Attention Module (CBAM) \cite{woo2018cbam}. 

This work studies the usage of Attention Mechanisms on DifferNet, a state-of-the-art anomaly detection method based on normalizing flows, by integrating attention blocks such as CBAM and SENet in its architecture. The new method is tested on three anomaly detection datasets: MVTec AD (public industrial dataset in controlled scenarios), the Semiconductor Wafer dataset \cite{schlosser2022improving} (private industrial dataset with real faulty data), and our InsPLAD-fault (under review/to be published, an in-the-wild dataset of power line asset inspection). The main contributions of this work are:

\begin{itemize}
    \item The new Attention-based DifferNet is superior to the standard DifferNet on all objects from three anomaly detection datasets, each dataset from a distinct domain;
    \item The Attention-based DifferNet achieves state-of-the-art performance on InsPLAD-fault (in the wild);
    \item The Attention-based DifferNet is qualitatively superior over standard DifferNet considering the most quantitatively improved categories;
    \item A straightforward coupling of attention mechanisms to modern feature-embedding-based unsupervised anomaly detection.
\end{itemize} 

\section{AttentDifferNet}



DifferNet is a state-of-the-art method for unsupervised image-based anomaly detection, combining convolutional neural networks with normalizing flows. The CNN in DifferNet is an AlexNet \cite{krizhevsky2017alexnet}, which works as a backbone for feature-embedding extraction. It takes the training images to generate descriptive features of flawless images. The features are then mapped to a latent space using a Normalizing Flow model. It is possible to calculate the likelihood of image samples from this latent space, and anomalous images should present a lower likelihood than the flawless samples present in the training process. Because of this, the training goal is to find parameters that maximize the likelihood of extracted features in the latent space.

DifferNet was conceived to detect defects in objects from images captured in a controlled context, such as objects from an industrial production line. To adapt it to overcome the challenges of object inspection in the wild, modular attention-based mechanisms were added to its backbone architecture. This allows the backbone network to focus on foreground elements and generate more relevant feature embeddings of the image with the inspected object. In this work, two architectures are experimented with, one using Squeeze-and-Excitation Networks \cite{hu2018senet} and the other using Convolutional Block Attention Modules \cite{woo2018cbam}.

SENets and CBAMs are well-known architectural unit attention mechanisms with similar objectives: to increase the representation power of CNNs by selectively emphasizing important features and suppressing irrelevant ones. While SENet focuses on modeling channel-wise relationships efficiently, CBAM infers intermediate attention maps along both channel and spatial dimensions to refine features adaptively.

\begin{figure}[t]
    \centering
    \includegraphics[width=\linewidth]{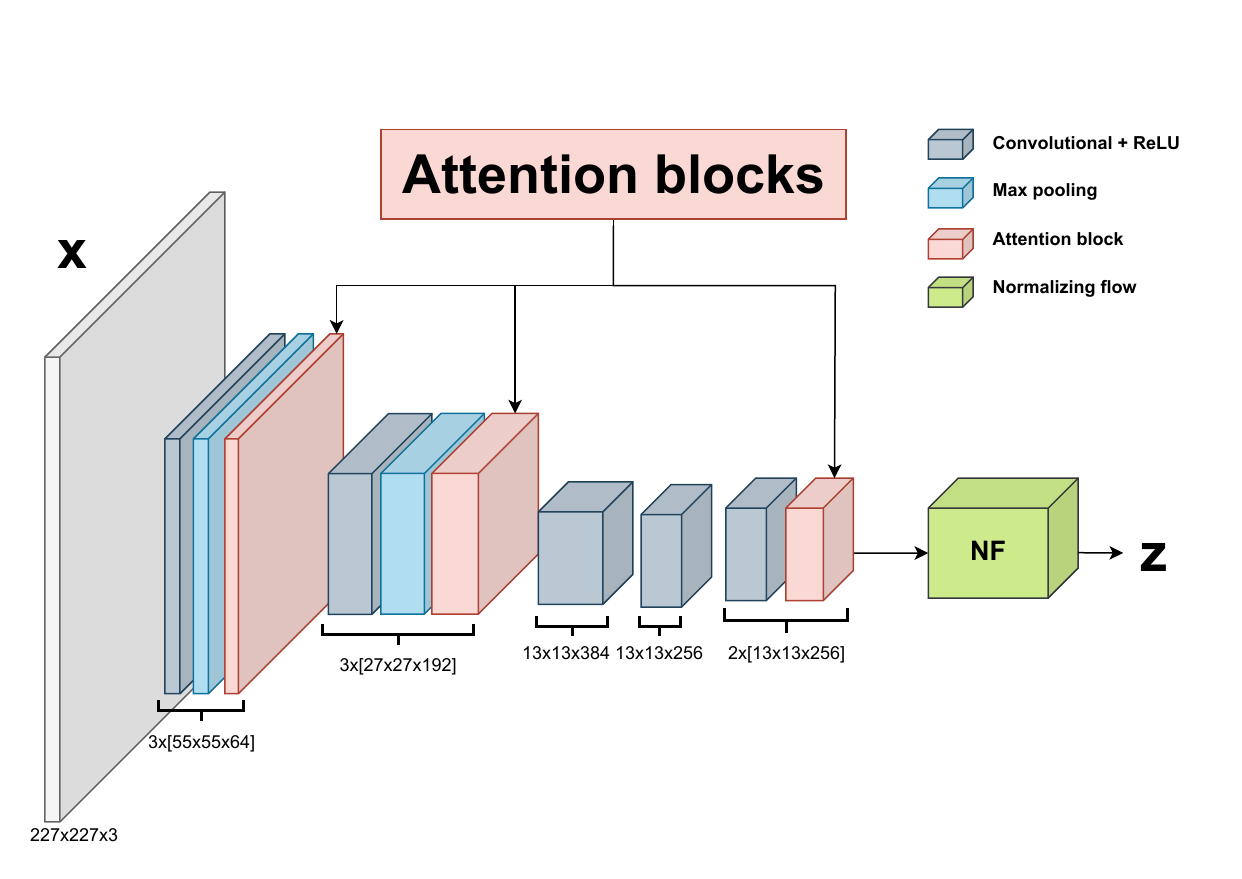}
    \caption{Proposed AttentDifferNet architecture.}
    \label{fig:attentdiffernet}
\end{figure}

Figure \ref{fig:attentdiffernet} shows our proposed architecture. The attention block's role changes according to the depth in which it is placed within the neural network. In the first few layers, it learns to highlight lower-level, class-agnostic features. In the deeper layers, it becomes more specialized, responding to different inputs in a class-specific manner. Therefore, our proposed architecture leverages the advantages of attention blocks throughout the entire network.

\section{Experiments}


For our experiments, we use three datasets which are briefly described below.

\begin{figure}[t]
    \centering

    \begin{subfigure}[b]{0.19\linewidth}
        \includegraphics[width=\linewidth]{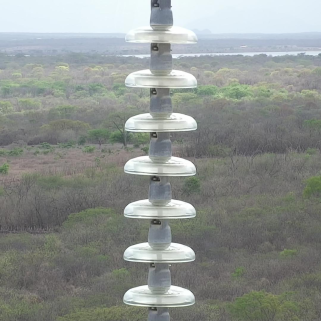} 
    \end{subfigure}
    \begin{subfigure}[b]{0.19\linewidth}
        \includegraphics[width=\linewidth]{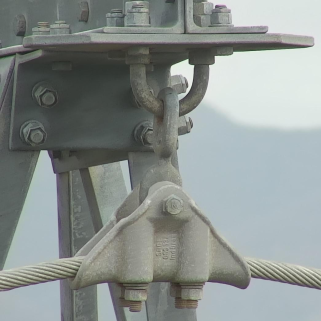} 
    \end{subfigure}
    \begin{subfigure}[b]{0.19\linewidth}
        \includegraphics[width=\linewidth]{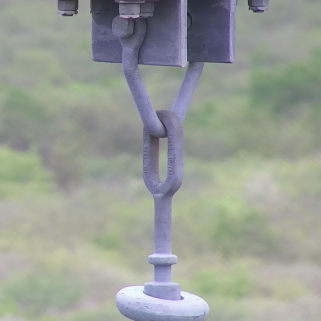} 
    \end{subfigure}
    \begin{subfigure}[b]{0.19\linewidth}
        \includegraphics[width=\linewidth]{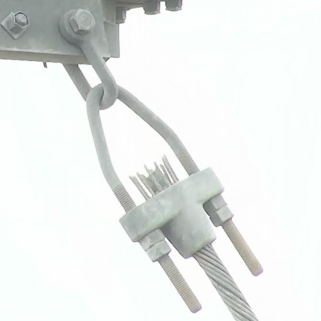} 
    \end{subfigure}
    \begin{subfigure}[b]{0.19\linewidth}
        \includegraphics[width=\linewidth]{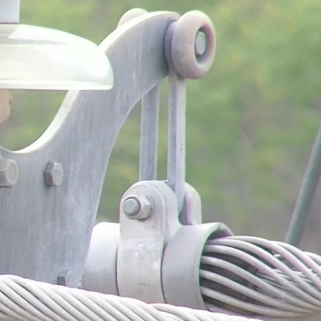}
    \end{subfigure}
    \begin{subfigure}[b]{0.19\linewidth}
        \includegraphics[width=\linewidth]{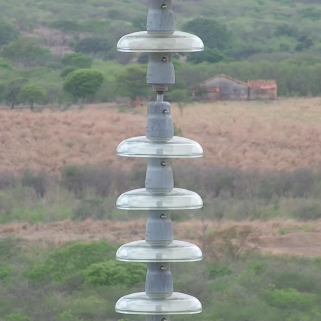} 
    \end{subfigure}
    \begin{subfigure}[b]{0.19\linewidth}
        \includegraphics[width=\linewidth]{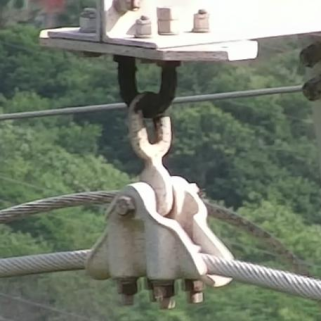} 
    \end{subfigure}
    \begin{subfigure}[b]{0.19\linewidth}
        \includegraphics[width=\linewidth]{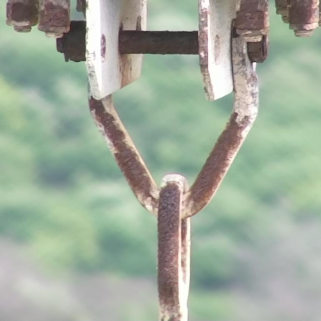} 
    \end{subfigure}
    \begin{subfigure}[b]{0.19\linewidth}
        \includegraphics[width=\linewidth]{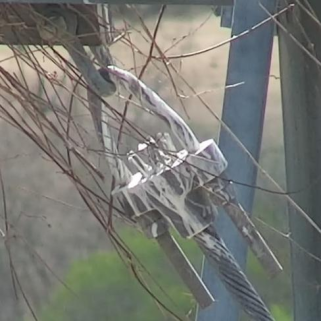} 
    \end{subfigure}
    \begin{subfigure}[b]{0.19\linewidth}
        \includegraphics[width=\linewidth]{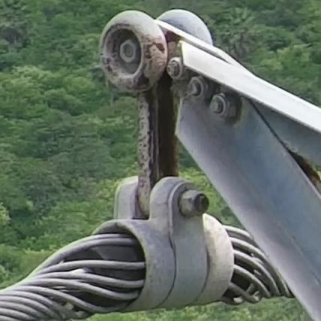}
    \end{subfigure}

    \caption{InsPLAD samples. The first row shows flawless assets, while the second shows defective ones. From left to right: Glass Insulator, Lightning Rod Suspension, Polymer Insulator Upper Shackle, Vari-grip, and Yoke Suspension.}
    \label{fig:insplad}
\end{figure}

\textbf{InsPLAD} is a power line asset inspection in-the-wild dataset that offers multiple computer vision challenges, one being anomaly detection in power line components called InsPLAD-fault. Its data are real-world Unmanned Aerial Vehicle (UAV) images of operating power line transmission towers. It contains five power line object categories with one or two types of anomalies for each class, resulting in $11\,662$ images, of which $402$ are samples of defective objects annotated on an image level. Since they are real-world defects, none of the faults have been fabricated or generated manually. Table~\ref{tab:inspladdes} shows the InsPLAD-fault properties for the anomaly detection task, whereas Figure~\ref{fig:insplad} depicts a flawless and defective sample for each of the five power line object classes. Its related publication is currently under review and the dataset will be made publicly available.

\begin{table}
  \centering
  \resizebox{\linewidth}{!}{%
\begin{tabular}{@{\extracolsep{\fill}}lccc@{\extracolsep{\fill}}}
\toprule
\multirow{4}{*}{Asset category} & \multicolumn{3}{c}{Anomaly detection}     \\ \cmidrule(lr){2-4}  
                                & \multicolumn{1}{c}{Train} & \multicolumn{2}{c}{Test}      \\ \cmidrule(lr){2-2} \cmidrule(lr){3-4}
                                & Flawless        & Flawless        & Anomalous         \\ 
\midrule
Glass Insulator             & 2298         & 581         & 90           \\
Lightning Rod Suspension    & 462          & 117         & 50           \\
\makecell[l]{Polymer Insulator Upper Shackle}    & 935          & 235         & 102          \\
Vari-grip                   & 477          & 114         & 63/48        \\ 
Yoke Suspension             & 4834         & 1207        & 49           \\
    \bottomrule
  \end{tabular}
  }
  \caption{InsPLAD-fault anomaly detection dataset description. Glass Insulator anomalies are missing caps, while the rest is corrosion-related. Vari-grip has two types: bird nest/corrosion.}
  \label{tab:inspladdes}
\end{table}

\textbf{MVTec AD} \cite{bergmann2019mvtec} is the most popular public dataset for unsupervised anomaly detection. It contains annotated data of objects and textures in controlled industrial scenarios at both image and pixel levels with and without anomalies. The anomalies are manually generated in an attempt to mimic real-world defects. It has ten objects and five textures categories, as shown in Table \ref{tab:mvtecad}. 

The \textbf{Semiconductor Wafer dataset} \cite{schlosser2022improving} is a visual inspection wafer dataset for image classification (annotated in image-level), encompassing various wafers, chips, streets, and street segments. Wafer images were obtained from various real-world dicing manufacturers by scanning the wafers’ chips after their cutting process. Different from MVTec AD, it contains real faulty data.




\subsection{Quantitative results}

Tables \ref{tab:insplad}, \ref{tab:mvtecad} and \ref{tab:semiconwafer} show the quantitative results for all three evaluated datasets. Our AttentDifferNets are compared to other state-of-the-art anomaly detection methods based on feature-embedding extraction. Values in bold font indicate the best result for a given category, while underlined values highlight the best result between the three DifferNet variations: standard DifferNet, AttentDifferNet (SENet), and AttentDifferNet (CBAM). 

Table \ref{tab:insplad} shows that the AttentDifferNet implementation that uses the SENet attention blocks achieved superior results in all categories of the InsPLAD-fault, not only compared to the DifferNet variations, but to all other state-of-the-art techniques. For the MVTec dataset, Table \ref{tab:mvtecad} shows a higher performance for FastFlow, achieving improved results in 73\% of the categories. The DifferNet using the CBAM attention block got the highest overall performance in one-third of the categories. However, it is worth mentioning that both AttentDifferNet variations outperformed the regular DifferNet in every category. Finally, on the Semiconductor Wafer dataset, CS-Flow has the higher overall average AUROC. When comparing DifferNet variations, it is noted that both AttentDifferNets outperformed the standard DifferNet.

The results indicate that using the attention blocks increased the average performance in the three datasets, with a highlight on AttentDifferNet's (SENet) performance on the InsPLAD-fault dataset, outperforming every other method tested with in-the-wild data. Another highlight is AttentDifferNet (CBAM) reaching the best AUROC in the Screw category of MVTec AD, surpassing all compared methods.

\begin{table*}[ht]
    \centering

    \resizebox{\textwidth}{!}{%
    \begin{tabular}{lcccccccccc}
    \toprule
    Category & DifferNet & \makecell{AttentDifferNet \\(SENet)} & \makecell{AttentDifferNet \\(CBAM)} & CS-Flow & PatchCore & Fastflow & CFLOW-AD \\
    \midrule
    Glass Insulator                 & 82,81\%       & \underline{\textbf{86,57\%}} & 81,03\%       & 85,73\% & 78,44\%       & 80,82\% & 82,22\% \\
    Lightning Rod Suspension        & 99,08\%       & \underline{\textbf{99,62\%}} & 99,33\% & 96,60\%       & 85,11\%       & 87,98\% & 95,52\% \\
    Polymer Insulator Upper Shackle & 92,42\% & \underline{\textbf{94,62\%}} & 92,10\%       & 88,40\%       & 81,02\%       & 87,57\% & 86,60\% \\
    Vari-Grip                       & 91,20\%       & \underline{\textbf{93,52\%}} & 88,99\%       & 91,53\%       & 91,92\% & 81,89\% & 90,37\% \\
    Yoke Suspension                 & 96,77\% & \underline{\textbf{97,38\%}} & 96,86\%       & 90,70\%       & 58,06\%       & 80,40\% & 83,87\% \\ \midrule
    Average AUROC                   & 92,46\% & \underline{\textbf{94,34\%}} & 91,66\%       & 90,59\%       & 78,91\%       & 83,73\% & 87,72\%\\
    \bottomrule
    \end{tabular}%
    }

    \caption{Comparison of AUROC results on the InsPLAD-fault dataset. Bold font indicates the best category result, while underlined values show the best result between our DifferNet variations.}
    \label{tab:insplad}
\end{table*}


\begin{table*}[ht]
  \centering
  \resizebox{\textwidth}{!}{%
    \begin{tabular}{lcccccccc}
    \toprule
    Category & DifferNet & \makecell{AttentDifferNet \\(SENet)} & \makecell{AttentDifferNet \\(CBAM)} & CS-Flow & PatchCore & Fastflow & CFLOW-AD \\
    \midrule
    Bottle        & 99,00\% & \underline{99,84\%}           & 99,68\%                 & 99,80\%          & \textbf{100\%} & \textbf{100\%}   & \textbf{100\%}    \\
Cable         & 95,90\% & \underline{98,43\%}           & 96,65\%                 & 99,10\%          & 99,50\%        & \textbf{100\%}   & 97,59\%           \\
Capsule       & 86,90\% & \underline{93,86\%}           & 92,58\%                 & 97,10\%          & 98,10\%        & \textbf{100\%}   & 97,68\%           \\
Carpet        & 92,90\% & 93,74\%                 & \underline{95,18\%}           & \textbf{100\%}   & 98,70\%        & \textbf{100\%}   & 98,73\%           \\
Grid          & 84,00\% & 90,89\%                 & \underline{91,23\%}           & 99,00\%          & 98,20\%        & \textbf{99,70\%} & 99,60\%           \\
Hazelnut      & 99,30\% & 99,89\%                 & \underline{\textbf{100,00\%}} & 99,60\%          & \textbf{100\%} & \textbf{100\%}   & 99,98\%           \\
Leather       & 97,10\% & 98,61\%                 & \underline{99,32\%}           & 100\%            & \textbf{100\%} & \textbf{100\%}   & \textbf{100,00\%} \\
Metal Nut     & 96,10\% & 96,53\%                 & \underline{97,70\%}           & 99,10\%          & \textbf{100\%} & \textbf{100\%}   & 99,26\%           \\
Pill          & 88,80\% & 91,79\%                 & \underline{93,48\%}           & 98,60\%          & 96,60\%        & \textbf{99,40\%} & 96,82\%           \\
Screw         & 96,30\% & 96,21\%                 & \underline{\textbf{98,93\%}}  & 97,60\%          & 98,10\%        & 97,80\%          & 91,89\%           \\
Tile          & 99,40\% & \underline{\textbf{100,00\%}} & \underline{\textbf{100,00\%}} & \textbf{100\%}   & 98,70\%        & \textbf{100\%}   & 99,88\%           \\
Toothbrush    & 98,60\% & \underline{\textbf{100,00\%}} & \underline{\textbf{100,00\%}} & 91,90\%          & \textbf{100\%} & 94,40\%          & 99,65\%           \\
Transistor    & 91,10\% & \underline{94,08\%}           & 93,92\%                 & 99,30\%          & \textbf{100\%} & 99,80\%          & 95,21\%           \\
Wood          & 99,80\% & 99,83\%                 & \underline{\textbf{100,00\%}} & \textbf{100\%}   & 99,20\%        & \textbf{100\%}   & 99,12\%           \\
Zipper        & 95,10\% & \underline{96,30\%}           & 95,88\%                 & \textbf{99,70\%} & 98,80\%        & 99,50\%          & 98,48\%           \\ \midrule
Avg. AUROC & 94,69\% & 96,67\%                 & \underline{96,97\%}           & 98,72\%          & 99,06\%        & \textbf{99,37\%} & 98,26\%  \\ 
    \bottomrule
        \end{tabular}
        }
  \caption{Comparison of AUROC results on MVTec AD dataset. Bold font indicates the best category result, while underlined values show the best result between DifferNet variations.}
        
        \label{tab:mvtecad}
\end{table*}

\begin{table*}[ht!]
  \centering
\resizebox{\textwidth}{!}{%
\begin{tabular}{lccccccc}
\toprule
Category & DifferNet & \makecell{AttentDifferNet \\(SENet)} & \makecell{AttentDifferNet \\(CBAM)} & CS-Flow & PatchCore & Fastflow & CFLOW-AD \\
\midrule
Street Classification & 86,40\%                       & \underline{90,44\%}                           & 84,53\%                                   & \textbf{97,19\%}            & 79,26\%                       & 80,94\%                      & 70,86\%                      \\
Chip Classification                   & 91,09\%                       & 89,96\%                                 & \underline{93,39\%}                             & 90,31\%                     & \textbf{93,90\%}              & 76,27\%                      & 92,01\%                      \\ \midrule
Average AUROC                                                                                   & 88,74\%                       & \underline{90,20\%}                           & 88,96\%                                   & \textbf{93,75\%}                     & 86,58\%                       & 78,60\%                      & 81,44\%                    \\\bottomrule
\end{tabular}
}
  \caption{Comparison of AUROC results on Semiconductor Wafer dataset. Bold font indicates the best category result, while underlined values show the best result between DifferNet variations.}

\label{tab:semiconwafer}
\end{table*}


\subsection{Qualitative results}

\begin{figure}
    \centering
    \setlength{\tabcolsep}{1pt}
    \begin{tabular}{cccc}
        \multicolumn{4}{c}{\textbf{DifferNet}} \\
        \noalign{\vskip 2pt}
        \includegraphics[width=0.24\columnwidth]{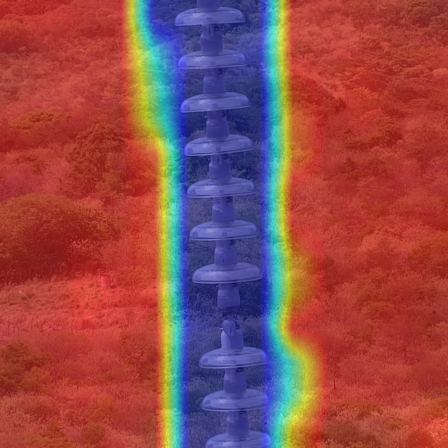}  &
        \includegraphics[width=0.24\columnwidth]{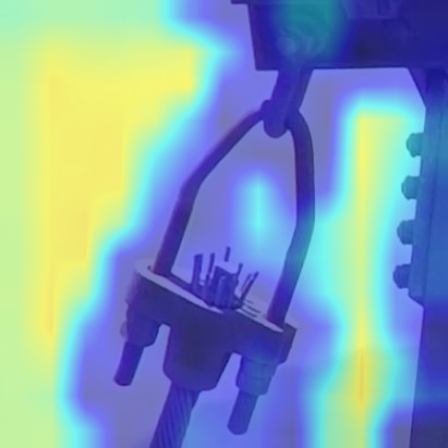} &
        \includegraphics[width=0.24\columnwidth]{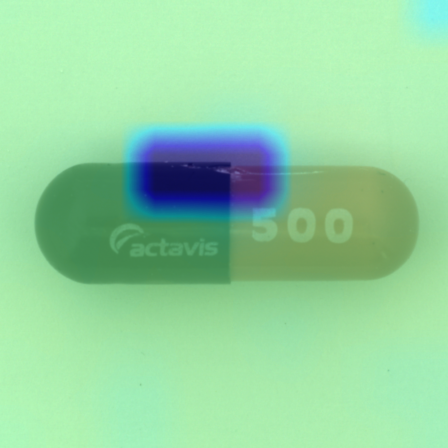} &
        \includegraphics[width=0.24\columnwidth]{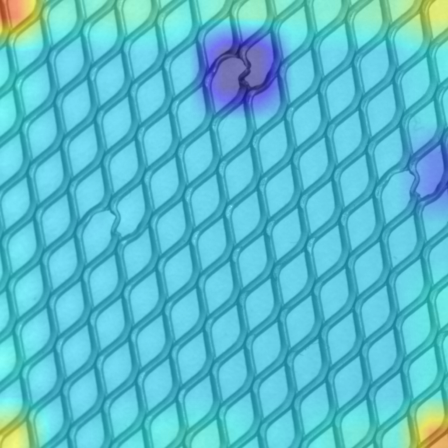} \\
        \noalign{\vskip 4pt}
        \multicolumn{4}{c}{\textbf{AttentDifferNet}} \\
        \noalign{\vskip 2pt}
        \includegraphics[width=0.24\columnwidth]{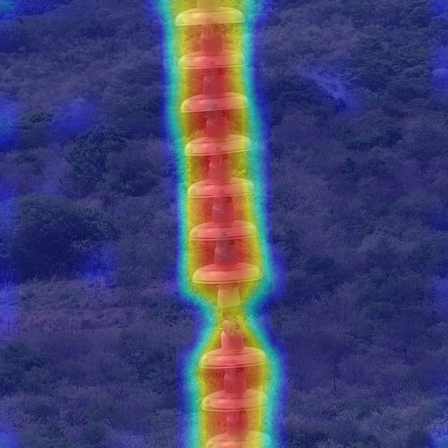}  &
        \includegraphics[width=0.24\columnwidth]{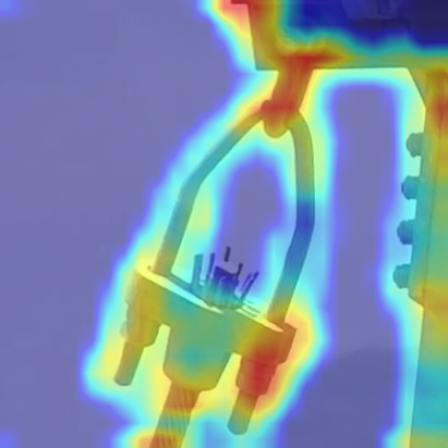} &
        \includegraphics[width=0.24\columnwidth]{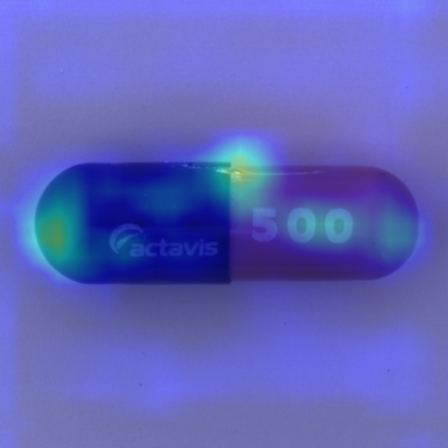} &
        \includegraphics[width=0.24\columnwidth]{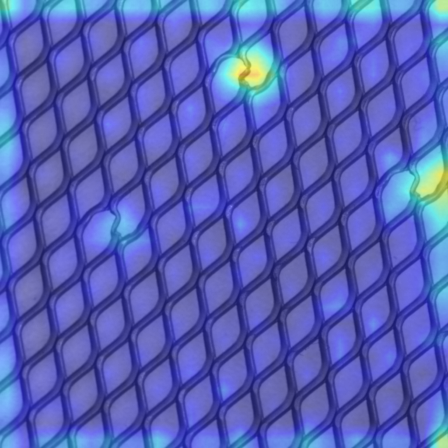} \\
    \end{tabular}

    \caption{Exemplary Grad-CAM-based class activation mapping comparison for DifferNet vs. AttentDifferNet given two categories from InsPLAD-fault (Glass Insulator and Vari-grip) and two from MVTec AD (Capsule and Grid), respectively.}
    \label{fig:quali}
\end{figure}

Here we show the qualitative results for some categories from two of the tested datasets. Figure \ref{fig:quali} compares the DifferNet and AttentDifferNet considering the two categories from InsPLAD-fault and two from MVTec AD. The Grad-CAM tool \cite{jacobgilpytorchcam,selvaraju2017gradcam} is used to reveal where the network is focusing to make its decisions. 
The two comparisons on the left side are from InsPLAD-fault in-the-wild data, and it is clear that AttentDifferNet focuses on the object and ignores the background. The Glass Insulator's missing cap is now taken into account.  
The two other comparisons are from MVTec AD data. AttentDifferNet was more specific on them, focusing on the anomaly itself, both from an object and a texture.

\section{Conclusion and outlook}

The main hypothesis of this work is that attention modules would help to improve the performance of state-of-the-art anomaly detection methods in an in-the-wild/uncontrolled environment scenario. We propose AttentDifferNet, an unsupervised anomaly detection method based on distribution mappings through normalizing flows that benefit from attention mechanisms by strategically coupling modular attention blocks to its feature extraction step. AttentDifferNet is able to achieve state-of-the-art performance on InsPLAD-fault, an anomaly detection in-the-wild dataset. We also show that AttentDifferNet not only maintains the model performance compared to DifferNet in controlled environments but it is able to improve it in virtually all categories of two relevant controlled environments' datasets for anomaly detection, the popular MVTec AD and the Semiconductor Wafer dataset.

This work implies that the state-of-the-art unsupervised anomaly detection methods have limitations in uncontrolled environments. It also portrays how the usage of attention blocks stands well suited to deal with such limitations.

\section*{Acknowledgements}


This research was co-funded by the German Academic Exchange Service (DAAD) and the Coordenação de Aperfeiçoamento de Pessoal de Nível Superior (CAPES) within the funding program “Co-financed Short-Term Research Grant Brazil, 2022 (ID: 57594818)”.

{\small
\bibliographystyle{ieee_fullname}
\bibliography{egbib}
}

\end{document}